\documentclass[runningheads]{llncs}
\usepackage{graphicx}
\graphicspath{{images/}}
\usepackage{amsmath,amssymb} 
\usepackage{color}
\usepackage{hyperref}

\usepackage[utf8]{inputenc}

\newcommand{\R}{{\mathbb{R}}}

\def\vec#1{\mathchoice%
	{\mbox{\boldmath $\displaystyle\bf#1$}}
	{\mbox{\boldmath $\textstyle\bf#1$}}
	{\mbox{\boldmath $\scriptstyle\bf#1$}}
	{\mbox{\boldmath $\scriptscriptstyle\bf#1$}}}

\def\mat#1{\mathchoice{\mbox{\boldmath$\displaystyle\tt#1$}}
	{\mbox{\boldmath$\textstyle\tt#1$}}
	{\mbox{\boldmath$\scriptstyle\tt#1$}}
	{\mbox{\boldmath$\scriptscriptstyle\tt#1$}}}

\DeclareMathOperator{\tr}{tr}
\DeclareMathOperator{\ovec}{vec}
\def\Eq{Eq. }

\begin{document}
\title{CvxPnPL: A Unified Convex Solution to the Absolute Pose Estimation Problem from Point and Line Correspondences\thanks{The authors would like to thank Jacopo Cavazza and all present at the Optimization Methods in Geometric Vision seminar at the 2019 NII Shonan Meetings, for their suggestions and insightful discussions.}}
\titlerunning{CvxPnPL}
%
\author{Sérgio Agostinho\inst{1}\orcidID{0000-0001-7008-1756} \and
João Gomes\inst{1}\orcidID{0000-0002-3524-5556} \and
Alessio Del Bue\inst{2}\orcidID{0000-0002-2262-4872}}
%
\authorrunning{S. Agostinho, J. Gomes and A. Del Bue}
%

\institute{
Instituto de Sistemas e Robótica\\
Instituto Superior Técnico, Universidade de Lisboa \\
Portugal\\
\email{sergio.agostinho@tecnico.ulisboa.pt,jpg@isr.tecnico.ulisboa.pt}\\
\and
Fondazione Istituto Italiano di Tecnologia\\
Genoa, Italy\\
\email{alessio.delbue@iit.it}\\
}
\maketitle              
\begin{abstract}
We present a new convex method to estimate 3D pose from mixed combinations of 2D-3D point and line correspondences, the\linebreak Perspective-n-Points-and-Lines problem (PnPL). We merge the contributions of each point and line into a unified Quadratic Constrained Quadratic Problem (QCQP) and then relax it into a Semi Definite Program (SDP) through Shor's relaxation. This makes it possible to gracefully handle mixed configurations of points and lines. Furthermore, the proposed relaxation allows us to recover a finite number of solutions under ambiguous configurations. In such cases, the 3D pose candidates are found by further enforcing geometric constraints on the solution space and then retrieving such poses from the intersections of multiple quadrics. Experiments provide results in line with the best performing state of the art methods while providing the flexibility of solving for an arbitrary number of points and lines.

\end{abstract}
\section{Introduction}

The problem of estimating the relative 3D pose between an object and a camera, given a number of 2D-3D correspondences, is well studied and it enabled a number of very successful applications in Robotics and Augmented Reality (AR) \cite{marchand2016pose}. The absolute pose problem is challenging because under certain geometric configurations of 2D-3D points and/or lines, there can exist more than a unique valid 3D pose. In such cases, all acceptable poses need to be retrieved and it is up to the user to rely on external information to disambiguate which one is correct. As such, it is understandable the few attempts were made to tackle the problem through convex optimization. The existence of a countable number of solutions implies that the problem is naturally non-convex. However, we show that with with our approach, it is possible to address the problem with convex optimization and to retrieve a finite number of solutions. The method makes use of point and line correspondences, to leverage collinearity and coplanarity constraints as in \cite{pnpml_minimal_ramalingam2011pose,zhou}. 
We formulate our optimization problem as a Quadratic Constrained Quadratic Program (QCQP), which we further relax into a Semi Definite Program (SDP) using Shor's relaxation \cite{nesterov}. We experimentally verify that our relaxation is tight if there is a unique valid pose, but we also claim that \emph{the rank of the obtained (relaxed) solution is equal to number of existing valid poses}. Our method is the first convex formulation to solve the Perspective-n-Points-and-Lines (PnPL) problem from 2D-3D correspondences, being able to recover up to 4 ambiguous poses, while only returning the number of solutions required. We modified the formulation from Zhou et al. \cite{zhou} to fully exploit all geometric information provided by the point and line correspondences and, lastly, we present a modification to Kukelova's et al. E3Q3 \cite{e3q3} to handle the further constrained case of the intersection of 6 quadrics with 3 unknowns.  Our experimental results are in line with the most accurate state-of-the-art methods, but we believe that the main contribution of this work comes from its theoretical findings.

\section{Related Work}
\label{sec:related-work}

The literature in pose estimation from 2D-3D correspondences is extensive and a comprehensive review is out of scope for this paper. 
Our method is designed for a central camera and non-minimal combinations of points and lines, despite being able to handle the minimal or planar points-only cases. For this reason we restrict the review to the following sub-topics.

\textbf{Perspective-n-Points.} 
The first approach to effectively estimate pose from 2D-3D correspondences was the DLT \cite{abdel1971dlt}. The DLT recovers both the camera instrinsics and pose and as such, tends to achieve worst accuracy when compared to methods which make use of the intrinsic information of the camera. However, it could scale to an arbitrary number of correspondences with only a linear increase in computational complexity. Subsequent approaches to solve the minimal \cite{dementhon1992exact,gao2003complete} and non-minimal \cite{pnp_pnl_ansar2003linear} problems all suffered from poor scaling for a large number of points. 
Schweighofer and Pinz \cite{schweighofer2008globally} proposed an O(n) convex method to solve the PnP problem for general cameras: it formulated a SDP problem around a quaternion rotation and it handled the planar case separately. EPnP \cite{pnp_lepetit2009epnp} 
is 
also  an O(n) method 
that relies on a parameterization based on four control points and a linearization step to simplify the optimization problem. 
 Subsequent works avoided the linearization step and tackled directly the polynomial problem \cite{pnp_dls,Zheng_2013_ICCV}, some targeting robustness and outlier rejection \cite{li2012robust,Ferraz_2014_CVPR}, others proposing a formulation for universal cameras \cite{pnp_kneip2014upnp}. Among all, OPnP \cite{Zheng_2013_ICCV} shows the most accurate results for the central camera case.

 \textbf{Perspective-n-Lines.} 
The first 
works addressed the PnL problem in the minimal 3 lines case \cite{chen1990pose,dhome1989determination}. The minimal problem was revisited recently by Xu et al. \cite{xu2017pose}, which observed that there could be a maximum of eight possible solutions. Ansar and Daniilidis \cite{pnp_pnl_ansar2003linear} proposed one of the first PnL methods, but it struggles to scale with 
large number of correspondences. The same year, Hartley and Zisserman \cite{hartley2003multiple} proposed an adaptation to the DLT for lines. 
Mirzae and Roumelioutis \cite{mirzaei2011globally} estimate the camera rotation matrix from a system of polynomial equations whose solution is extracted from an eigen-decomposition. 
\text{P{\v{r}}ibyl} et al. \cite{pvribyl2015camera,pvribyl2017absolute} proposed a DLT inspired method which makes use of Pl\"{u}cker line parameterization to formulate the problem as a system of linear equations. 
Zhang et al. \cite{zhang2012robust} proposed a robust method to estimate the pose from multiple line triplets. Lastly, Zhou et al. \cite{zhou2019robust} addressed the PnL problem in terms of two algebraic distances to approximate Geometric distance. The method skips the use of a Gr\"{o}bner basis solver and solves the first order optimality conditions of its polynomial equation using a more stable hidden variable method.

\textbf{Perspective-n-Points-and-Lines.} The literature for mixed combinations of points and lines is briefer compared to the previous two modalities. The works of Ramaligan et al. \cite{pnpml_minimal_ramalingam2011pose} and Zhou et al. \cite{zhou} address the minimal cases while for non-minimal cases the DLT approach \cite{hartley2003multiple} can naturally be adapted to take contribution from points and lines. Subsequently, Kuang and {\AA}str{\"{o}}m \cite{Kuang_2013_ICCV} proposed a method to jointly estimate the pose and focal length from points, lines and points with direction. Vakhitov et al. \cite{vakhitov2016accurate} proposed an adaptation to EPnP and OPnP and extended their support to lines.

\textbf{Convex Relaxations on Rotation Matrices.} Saunderson et al. \cite{saunderson2015semidefinite} provided an in-depth analysis on the properties of the convex hull enclosing rotation matrices and the works \cite{rosen2015convex,khoo2016non,Briales_2017_CVPR,Briales_2018_CVPR,rosen2019se,yang2019polynomial} showed that despite the non-convex nature of the SO(3) group, there exist successful relaxation strategies. Carlone et al. \cite{carlone2016planar} solved the planar pose graph optimization problem resorting to an SDP relaxation. This relaxation is mostly tight and in the cases where it is not, it still provided a suboptimal reduced search space to a extract a meaningful solutions. This work was later extended to the full 3D case by Rosen et al \cite{rosen2019se}. Our method draws inspiration from the recent work of Briales et al. in \cite{Briales_2017_CVPR,Briales_2018_CVPR} that show: the possibility to enforce the orthogonal and the determinant constraints of a rotation matrix using only quadratic constraints, apply a QCQP relaxation and achieve remarkable results for non-minimal 3D registration and relative pose between two views problems. 

\section{A Unified Formulation for Point and Line Correspondences}
\label{sec:homo-formulation}

\begin{figure}[t]
\begin{center}
\includegraphics[width=\linewidth]{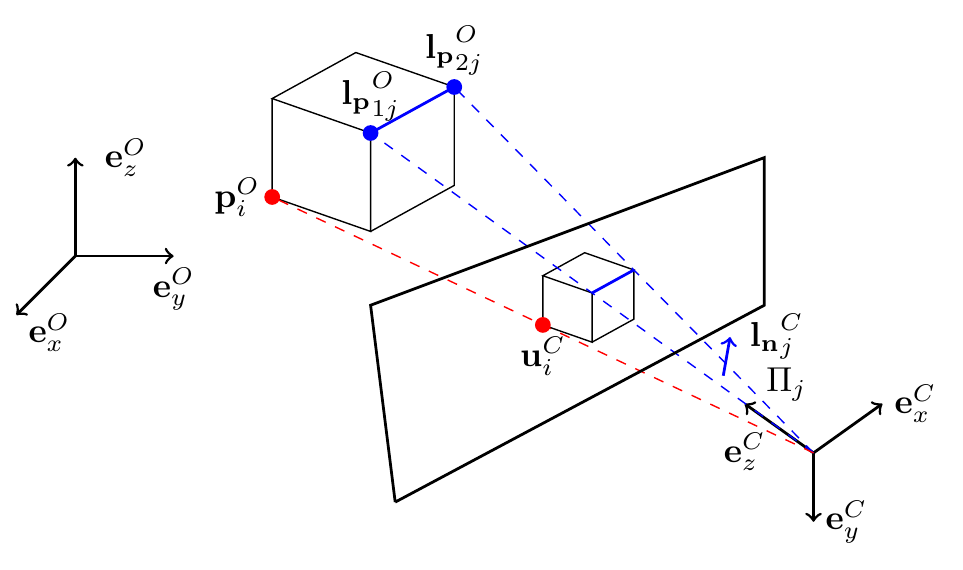}
\end{center}
\caption{Example of 2D-3D correspondences from a single point and line. Elements belonging to point correspondences are represented in red and elements belonging to line correspondences are represented in blue. The superscripts $O$ and $C$ denote that the element belongs to the object and camera's reference frames, respectively.}
\label{fig:correspondences}
\end{figure}

We are interested in formulating an optimization problem, with respect to the 3D pose of a model, which combines geometric information from both points and lines. To do so, we employ the collinear and coplanarity constraints, introduced by Ramalingam et al. \cite{pnpml_minimal_ramalingam2011pose}. Let us consider $\vec{p}_i$ as a 3D point from a set of $n$ points, and the tuple $(\vec{l_p}_{1j}, \vec{l_p}_{2j})$ as two points parameterizing a 3D line from a set of $m$ lines, defined in the object's reference frame. Refer to Figure \ref{fig:correspondences} to aid in the visualization of the problem geometry. The 3D model's pose with respect to the camera is given by the rotation matrix $\mat{R}$ and the translation vector $\vec{t} \in \mathbb{R}^3$. We also define $\mathbf{r}$ as the vectorization of $\mat{R}$ such that $\vec{r} = \ovec(\mat{R})$.

\textbf{Point Correspondences.}
The projection of a 3D point $\vec{p}_i$ onto the image plane will land on a given
pixel and the back projection of this pixel into the camera's 3D space
will form the ray $\vec{u}_i$. Both $\vec{p}_i$ and $\vec{u}_i$ are necessarily collinear, allowing us to enforce $\vec{u}_i \times \left(\mat{R} \vec{p}_i + \vec{t}\right) = 0$, where the operator $\times$ represents the cross product. Then, let us  write $\lfloor \mathbf{u}_i \rfloor_\times \left(\mat{R} \mathbf{p}_i + \mathbf{t}\right) = 0$, employing the equivalent skew symmetric matrix representation $\lfloor \mathbf{u} \rfloor_\times$ for the cross product. 
The matrix $\lfloor \mathbf{u} \rfloor_\times$ has rank 2, so despite each point correspondence contributing 3 equations, only 2 of them are linearly independent. Stacking the contributions from all $n$ points and rearranging terms yields $\mat{C_p} \vec{r} + \mat{N_p} \vec{t} = 0$,
where $\mat{C_p}$ and $\mat{N_p}$ are $3n \times 9$ and $3n \times 3$ matrices, respectively \footnote{\label{foot:supp} The full derivation of this expression can be found in the supplemental material.}.

\textbf{Line Correspondences.}
Any 3D line, which is not collinear with the origin of the camera's reference frame, forms a unique plane $\Pi_j$ with it (see Figure \ref{fig:correspondences}). We denote by $\vec{l_n}_j$ the normal of $\Pi_j$. The line constraints are built from the fact that in the camera's space, both $\vec{l_p}_{1j}$ and $\vec{l_p}_{2j}$ belong to $\Pi_j$ and are therefore orthogonal to $\vec{l_n}_j$, respecting
\begin{eqnarray}
\vec{l_n}_j \cdot \left(\mat{R} \vec{l_p}_{ij} + \vec{t}\right) &=& 0 \quad : i = {1, 2}.
\end{eqnarray}
Each line correspondence equally contributes to two linearly independent equations and upon stacking, it allows us to form the system
\begin{equation}
\mat{C_L} \vec{r} + \mat{N_L} \vec{t} = 0,
\end{equation}
where $\mat{C_L}$ is a  $2m\times 9$ matrix, and $\mat{N_L}$, $2m\times 3$ \textsuperscript{\ref{foot:supp}}. Consider $\vec{u_l}_{j1}$ and $\vec{u_l}_{j2}$, the bearing vectors associated with two distinct points sampled from the 2D line projection. The normal $\vec{l_n}_j$ can be recovered from their cross product $\vec{l_n}_j = \lfloor \vec{u_l}_{j1} \rfloor_\times \vec{u_l}_{j2}$.

\subsection{Composing the Homogeneous System}

Stacking together the contributions from  points and lines requires us to further define the matrices as:
\begin{eqnarray}
\mat{C} &=& \left[\begin{matrix}\mat{C_p} \\ \mat{C_L}\end{matrix}\right] \\
\mat{N} &=& \left[\begin{matrix}\mat{N_p} \\ \mat{N_L}\end{matrix}\right].
\end{eqnarray}
Both $\mat{C}$ and $\mat{N}$ are composed of $3n + 2m$ rows, where only $2n + 2m$ are linearly independent. It is a known linear algebra result, that given an optimal $\vec{\hat{r}}$, the optimal unconstrained solution $\vec{\hat{t}}$ to the overdetermined system
\begin{equation}
\mat{C} \mathbf{r} + \mat{N} \mathbf{t} = 0,\label{eq:linear-rt}
\end{equation}
is given by $\vec{\hat{t}} = - (\mat{N}^T \mat{N})^{-1} \mat{N}^\top \mat{C} \vec{\hat{r}}$ and by
substituting back into Eq. \eqref{eq:linear-rt}, allows us to write the complete system as $\mat{A} \vec{r} = 0$, with
\begin{equation}
    \mat{A} = (\mat{I}_{3n + 2m} - \mat{N}(\mat{N}^\top \mat{N})^{-1} \mat{N}^\top)\mat{C},
\end{equation}
where $\mat{A}$ is a $(3n+2m)\times 9$ matrix and $\mat{I}_{3n + m}$ is the identity matrix of size $3n + 2m$. The second row block of $\mat{A}$, is the residual of the projection of $\mat{C}$ onto the column space of $\mat{N}$.

\section{Convex Formulation}
\label{sec:cvx-formulation}

The formulation in Eq. \eqref{eq:linear-rt} does consider the specific structure of $\vec r = \ovec(\mat{R})$. Proper rotation matrices are orthogonal and satisfy $\det(\mat{R}) = +1$. While the orthogonality is a quadratic constraint, the determinant is cubic. However, it was show in \cite{tron}, that enforcing the right-hand convention in the columns or rows, ensures a positive determinant, allowing us to express this inherently cubic constraint in an equivalent quadratic form. As such, we formulate our problem as:
\begin{eqnarray}
\operatorname*{min}_{\mathbf{r}} & & \|\mat{A} \mathbf{r}\|^2 \label{eq:cost} \\
\operatorname{s.t.\,} & & \mat{R}^\top \mat{R} = \mat{I}_3 \label{eq:constraint-cols}\\
& & \mat{R} \mat{R}^\top = \mat{I}_3 \label{eq:constraint-rows}\\
& & \mat{R}^{(i)} \times \mat{R}^{(j)} = \mat{R}^{(k)} \, : (i, j, k) \in \{(1, 2, 3), (2, 3, 1), (3, 1, 2)\}, \label{eq:constraint-right}
\end{eqnarray}
with $\mat{R}^{(i)}$ denoting the $i$-th column of $\mat{R}$. The previous problem can be written in a  canonical form given by \textsuperscript{\ref{foot:supp}}:
\begin{eqnarray}
\operatorname*{min}_{\mathbf{\tilde{r}}} & & \mathbf{\tilde{r}}^T \mat{Q}_0 \mathbf{\tilde{r}}  \\
\operatorname{s.t.\,} & & \mathbf{\tilde{r}}^T \mat{Q}_i \mathbf{\tilde{r}} = 0 : i \in \{1, \dots, 21\}  \label{eq:quad-constraints} \\
& & \mathbf{\tilde{r}}_{10} = 1,
\end{eqnarray}
where $\vec{\tilde{r}}$ is the homogeneous vector $\vec{\tilde{r}} = [\mathbf{r}^\top 1]^\top$ of size 10. It is also important to highlight that $\mat{Q_0}$ and $\mat{Q_i}$ belong to the space of symmetric matrices $\mathbb{S}_{10}$. The constraints in Eq. \eqref{eq:quad-constraints} are non-convex, so in order to overcome this limitation, we relax the problem into a SDP employing Shor's relaxation \cite{nesterov,shor}. This relaxation exploits the trace identity $\tr(\vec{\tilde{r}}^\top \mat{Q} \vec{\tilde{r}}) = \tr(\mat{Q} \vec{\tilde{r}} \vec{\tilde{r}}^\top)$,
allowing to rewrite the QCQP as:
\begin{eqnarray}
\operatorname*{min}_{\mat{Z}} & & \tr(\mat{Q}_0 \mat{Z})  \\
\operatorname{s.t.\,} & & \tr(\mat{Q}_i \mat{Z}) = 0 : i \in \{1, \dots, 21\} \\
& & \mat{Z} \succcurlyeq 0 \\
& & \mat{Z}_{10,10} = 1,
\end{eqnarray}
where $\mat{Z} = \vec{\tilde{r}} \vec{\tilde{r}}^\top$ is a $10 \times 10$ matrix with rank 1 by construction. The rank constraint is not convex and therefore it is dropped, relaxing the original problem. As it turns out, without this relaxation it is unlikely that we would be able to retrieve meaningful solutions in situations where there are more than one, as discussed in Section \ref{sec:solution-recovery}.

\section{Rotation Recovery}
\label{sec:solution-recovery}

The optimal $\vec{\tilde{r}}$ needs to verify the following structure:
\begin{eqnarray}
\tilde{\vec{r}} &=& \sum_{k=1}^K \alpha_k \vec{v}_{\lambda_k} \\
\tilde{r}_{10} &=& 1, \label{eq:sol-linear-constrt}
\end{eqnarray}
where $K$ denotes the rank of $\hat{\mat Z}$ and $\vec v_k$ are the eigenvectors associated with non-null eigenvalues. We empirically observed that \emph{the rank of $\hat{\mat Z}$ equals the number of ambiguous poses the problem has}. For additional insights on how this claim was verified please refer to Section 5 of the supplementary material. We refer as ambiguous poses, the candidate poses which still produce a minimum in \Eq \eqref{eq:cost}. For instance, the minimal P3P problem is known to have in general 4 ambiguous poses. This means that if one projects the 3D points to the image plane according to each of the 4 poses, all of them will produce the same 2D coordinates. Despite the existence of multiple poses, only one is meaningful. The rank is a proxy for the number of admissible solutions and it affects the complexity involved in the recovery of $\vec{\tilde{r}}$. While dealing with rank 1 is relatively straightforward, rank 2 and above requires enforcing geometric constraints as in Eqs. \eqref{eq:constraint-cols}, \eqref{eq:constraint-rows} and \eqref{eq:constraint-right} on the solution space in order to retrieve all admissible solutions.
In theory there  should not exist situations in which $\operatorname{rank}(\hat{\mat Z})$  is 3 as the geometry of the pinhole camera model provides a number of ambiguous poses as integer powers of 2 \cite{xu2017pose}. However, we experimentally observed that such cases can sporadically occur as a byproduct of an early stop by the solver e.g., because it reached the maximum number of allowed iterations.
Therefore, if $\operatorname{rank}(\hat{\mat Z})$  is 3, we treat such case as rank 4.
Enforcing the linear constraint in Eq. \eqref{eq:sol-linear-constrt} allows removing one of the unknowns resulting in an expression of the form:
\begin{eqnarray}
\tilde{\mathbf r} &=& \sum_{k=1}^{K-1} \alpha_k' \mathbf v_k' + \mathbf v_0' \label{eq:vec-sum} \\
\left[\begin{matrix} \mathbf r \\ 1\end{matrix}\right] &=& \sum_{k=1}^{K-1} \alpha_k' \left[\begin{matrix} \vdots \\ 0\end{matrix}\right] + \left[\begin{matrix} \vdots \\ 1\end{matrix}\right].
\end{eqnarray}
Imposing the rotation matrix constraints here results in a system of 21 quadratic polynomial equations with $K-1$ unknowns. This paper provides solutions up to rank 4,  
meaning that we are able to tackle points-only configurations with 3 or more points, but for lines-only or mixed configurations of points and lines, we require at least 4 elements to get a solution. The next paragraphs, and with more details in the supplemental material, provide the procedure to recover such solutions.

\textbf{Rank 1.}
\label{sec:rank-1}
When dealing with rank 1 matrix $\hat{\mat{Z}}$, the solution to the problem is unique and the relaxation is indeed tight. The optimal $\vec{\hat{r}}$ can be recovered from the eigenvector associated with the largest eigenvalue as $\vec{v}_{\lambda_{\text{max}}} = a \left[\begin{matrix}\vec{\hat{r}}^\top & 1\end{matrix}\right]^\top$.
In practice, it is also advisable to reproject this solution to the orthogonal matrix space, so after applying $\ovec^{-1}$, we decompose it using SVD and retrieve $\hat{\mat{R}}$ as:
\begin{eqnarray}
\mat{R}' &=& \ovec^{-1}(\vec{\hat{r}}) \\
\mat{U} \mat{D} \mat{V}^\top &=& \operatorname{svd}(\mat{R}') \\
\hat{\mat{R}} &=& \mat{U} \mat{V}^\top.
\end{eqnarray}

\textbf{Rank 2.}
\label{sec:rank-2}
Refer to Eq. \eqref{eq:vec-sum} with $K = 2$. After enforcing the linear constraint on the last element of $\tilde{\vec r}$, we have a single unknown, designated as $a$. After enforcing the quadratic constraints on the rotation we end up with a system of equations of the form $\mat G \left[\begin{matrix}a^2 & a & 1\end{matrix}\right]^\top = 0$\textsuperscript{\ref{foot:supp}},
where $\mat G \in \R^{21\times3}$. We empirically verify that $\mat G$ has rank 1, respecting our claim that the rank is indicative of the number of ambiguous solutions. We retrieve both values of $a$ from the row-wise average of $\mat G$, denoted as $\bar{\vec g}^\top$, by finding the roots of the second order polynomial $\bar{g}_1 a^2 + \bar{g}_2 a + \bar{g}_3 = 0$.
Because these roots are required to exist in the admissible solution space formed by the non-null eigenvectors of $\mat Z$, \emph{they are always real}.

\textbf{Rank 4.}
\label{sec:rank-4}
In the rank 4 case, we will designate the three unknown $\alpha_1'$, $\alpha_2'$ and $\alpha_3$ by the letters $a$, $b$ and $c$. After enforcing the available constraints we obtain the following system $\mat G \left[\begin{matrix}a^2 & b^2 & c^2 & ab & ac & bc & a & b & c & 1\end{matrix}\right]^\top = 0$,
where $\mat G \in \R^{21\times10}$. We empirically observed the $\mat G$ is composed of 6 linearly independent equations. This is equivalent to finding the intersections of 6 quadrics with 3 unknowns. 
Consider the column-wise block representation of $\mat G$ such that $\mat G = \left[\begin{matrix}\mat{G_L} & \mat{G_R}\end{matrix}\right]$,
with matrices $\mat{G_L} \in \R^{21\times6}$ and $\mat{G_R} \in \R^{21\times4}$.
Given the rank 4 of $\mat G$ and resorting to the left pseudo inverse, we can write
\begin{equation}
\left[\begin{matrix}a^2 \\ b^2 \\ c^2 \\ ab \\ ac \\ bc\end{matrix}\right] = \underbrace{(\mat{G_L}^\top \mat{G_L})^{-1} \mat{G_L}^\top \mat{G_R}}_{\mat D} \left[\begin{matrix} a \\ b \\ c \\ 1\end{matrix}\right].\label{eq:quartic-system}
\end{equation}
To find our solution we adapt the E3Q3 method developed by Kukelova et al. in \cite{e3q3}. We pick a variation rows 2, 3 and 6 and treat $a$ as a constant. Doing so, allows us to write Eq. \eqref{eq:quartic-system} as:
\begin{equation}
\left[\begin{matrix}b^2 \\ c^2 \\ bc\end{matrix}\right] =
\left[\begin{matrix}
d_{22} & d_{23} & d_{21} a + d_{24} \\
d_{32} & d_{33} & d_{31} a + d_{34} \\
d_{62} & d_{63} & d_{61} a + d_{64}
\end{matrix}\right]
\left[\begin{matrix}b \\ c \\ 1\end{matrix}\right]. \label{eq:coeff-selection}
\end{equation}
After applying the identities $(b^2)c = (bc)b$, $(c^2)b = (bc)c$ and $(b^2)(c^2) = (bc)(bc)$, followed by double substitution yields the homogeneous system \textsuperscript{\ref{foot:supp}}
\begin{equation}
\underbrace{\left[\begin{matrix}
m^{[1]}_{11}(a) & m^{[1]}_{12}(a) & m^{[1]}_{13}(a) \\
m^{[1]}_{21}(a) & m^{[1]}_{22}(a) & m^{[1]}_{23}(a) \\
m^{[1]}_{31}(a) & m^{[1]}_{32}(a) & m^{[2]}_{33}(a)
\end{matrix}\right]}_{M(a)}
\left[\begin{matrix}b \\ c \\ 1\end{matrix}\right] = 0 \label{qe:poly-homo}.
\end{equation}
The subscript $[\cdot]$ denotes the degree of the polynomial in $a$. Eq. \eqref{qe:poly-homo} only has a non-trivial solution if the determinant of $M(a)$ is 0. This amounts to finding the roots of a 4th degree polynomial, yielding our 4 desired solutions. 
Recovering $b$ and $c$ amounts to substituting $a$ in Eq. \eqref{qe:poly-homo} for every solution and solving the overdetermined linear system for $b$ and $c$. As the reader might notice, our particular selection of coefficients $((b,c),a)$ in Eq. \eqref{eq:coeff-selection} or rows in Eqs. \eqref{eq:quartic-system} is not unique. The same procedure can be done with a different row and coefficient selection.

\section{Experimental Results}
\label{sec:results}

We present separate results for the angular and translations errors when solving for the PnPL problem. The error metrics used are given by
\begin{eqnarray}
\Delta \mat{R} &=& \hat{\mat{R}}^\top \mat{R}_{gt}\\
\Delta \vec{t} &=& \tfrac{\|\hat{\vec{t}} - \vec{t}_{gt}\|}{\|\vec{t}_{gt}\|},
\end{eqnarray}
for rotation and translation errors respectively, with the subscript $gt$ denoting the ground truth. Given the residual rotation $\Delta \mat{R}$, the angular error is retrieved from the absolute value of the angle, once $\Delta \mat{R}$ is converted to its axis-angle representation. The translation error is computed in its normalized form, to prevent  situations where the object's origin $\vec{t}_{gt}$ is located far away from the camera origin, completely tainting and dominating the translation error statistics.

\begin{figure}
\begin{center}
\includegraphics[width=0.98\linewidth]{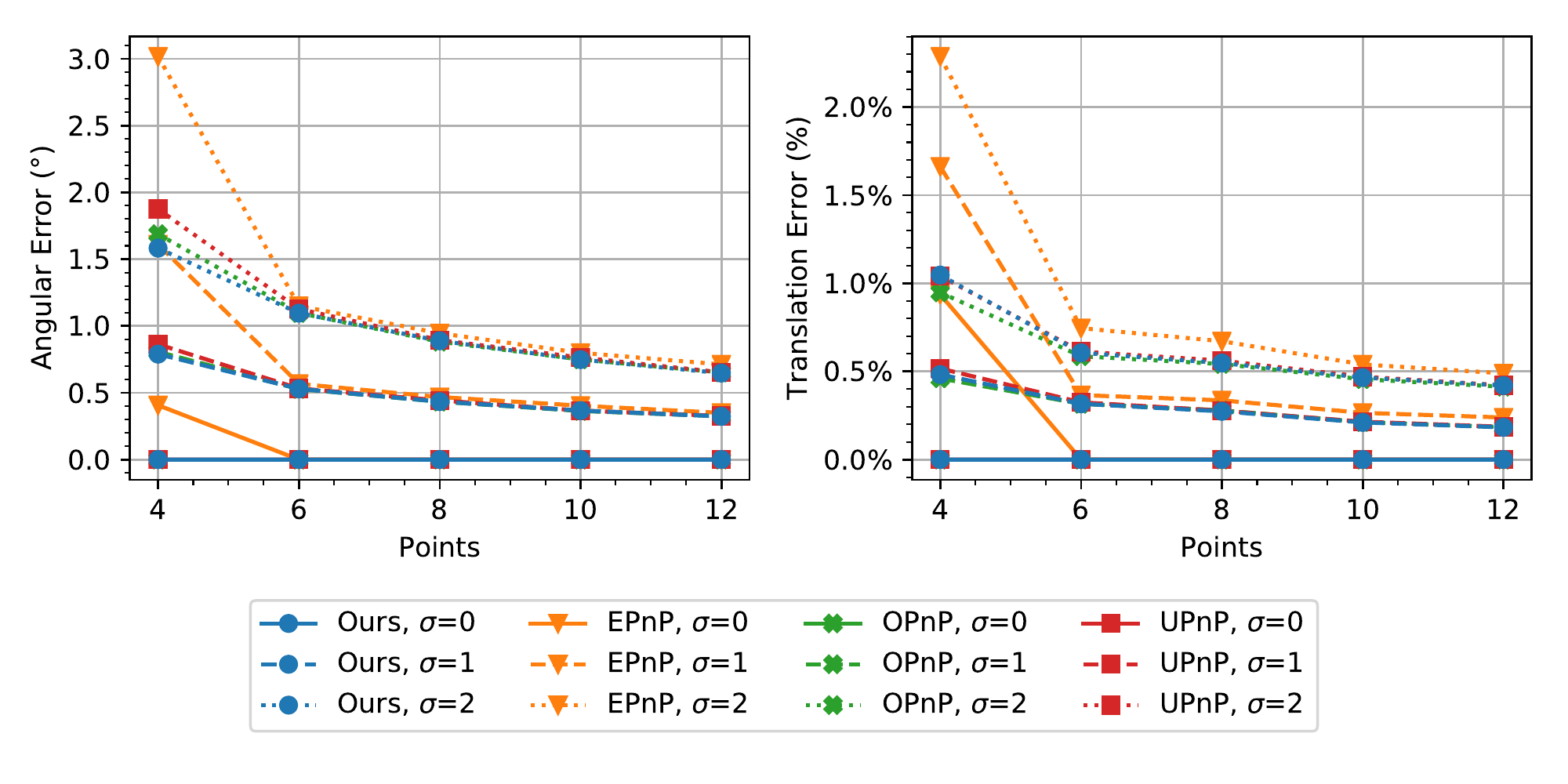}
\includegraphics[width=0.98\linewidth]{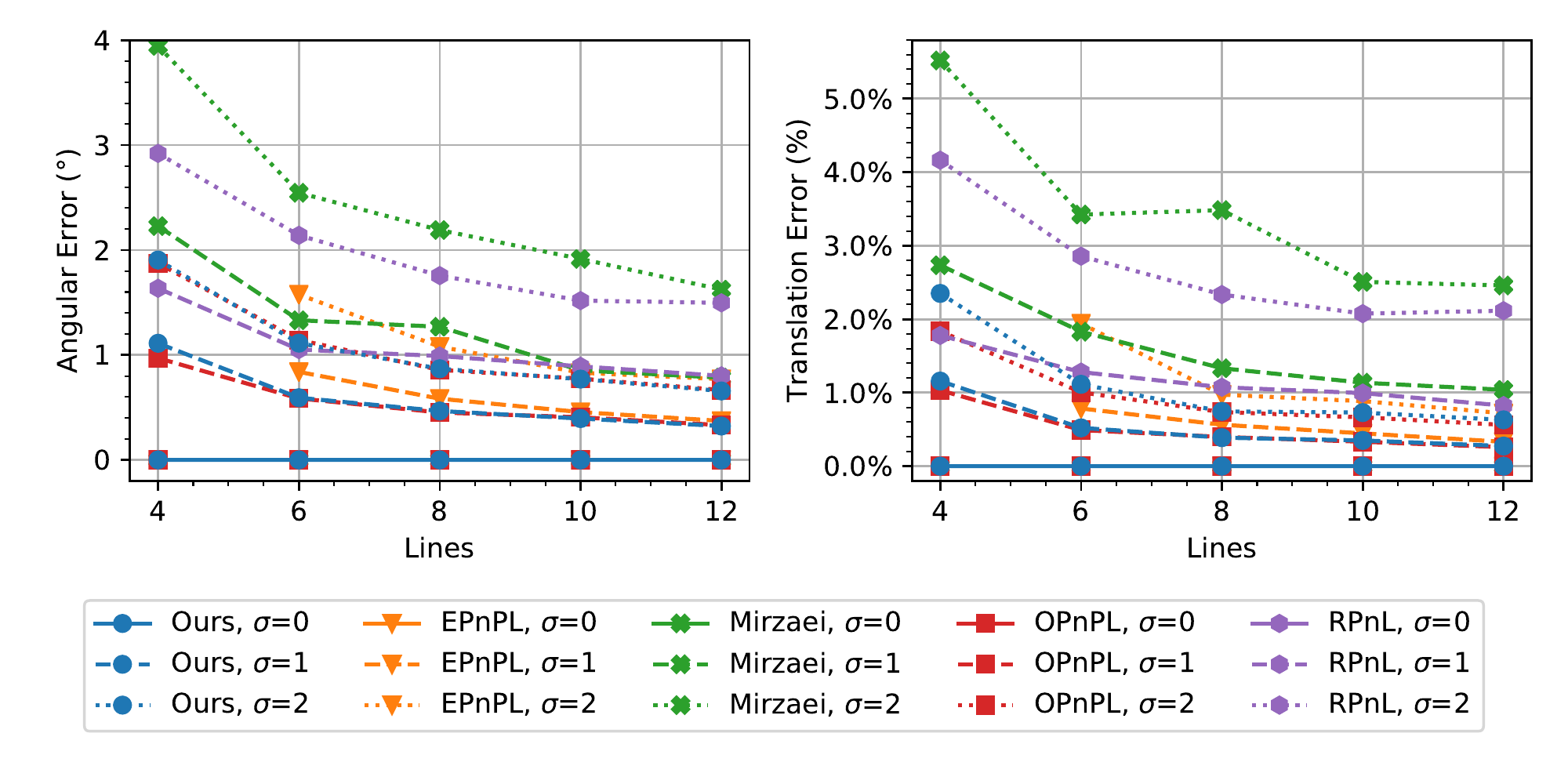}
\includegraphics[width=0.98\linewidth]{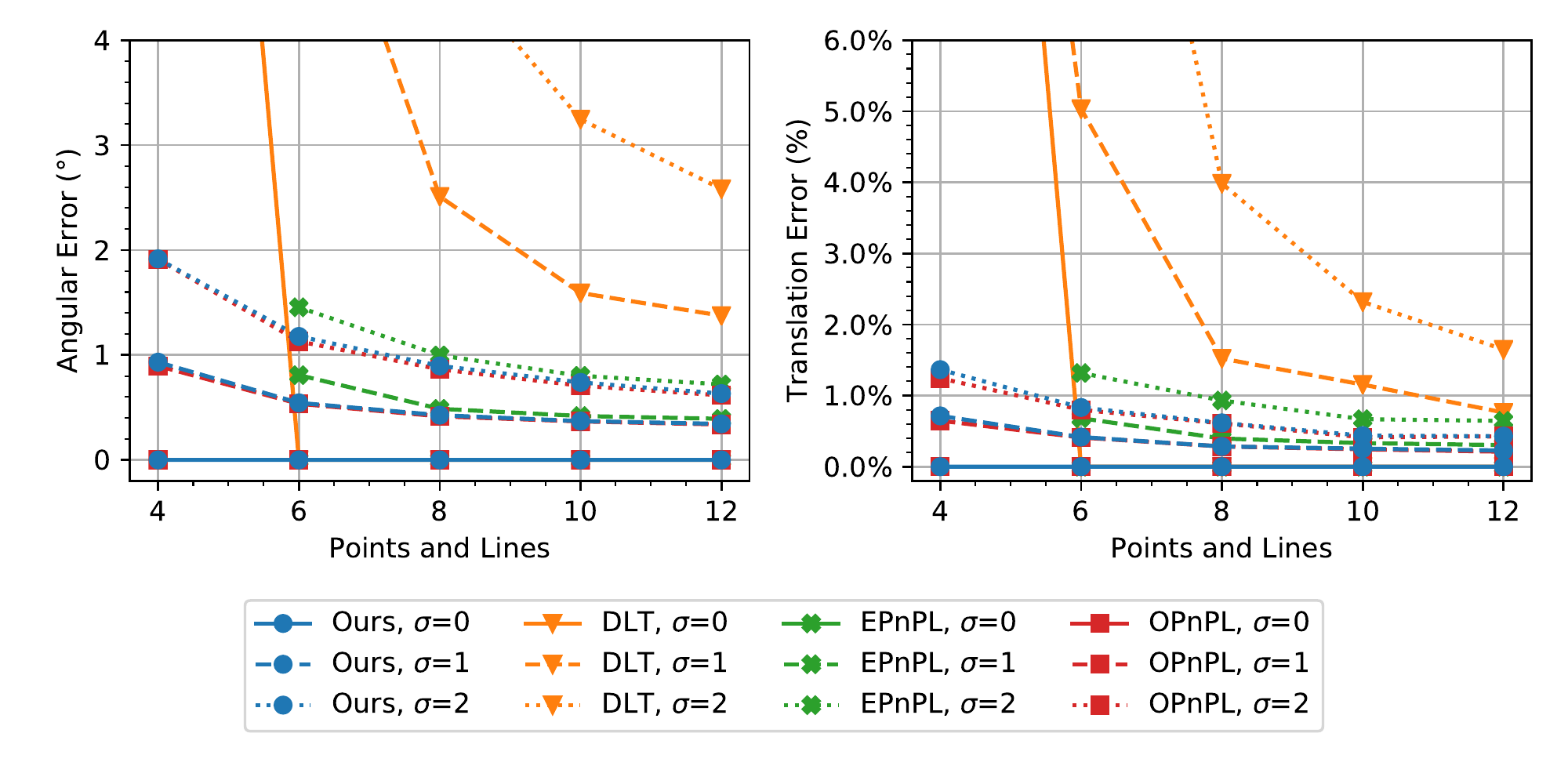}
\end{center}
   \caption{Comparison between different methods for a PnP scenario with 4+ points (top), PnL scenario with 4+ lines (middle) and a PnPL scenario with 4+ points and lines (bottom). The median angular and translation error are shown on the left and right, respectively, for different numbers of elements in the scenario, employing a Gaussian pixel noise with standard deviation $\sigma$ computed over $1000$ runs (figure best seen in colour).}
\label{fig:results-synth}
\end{figure}

\textbf{Simulation Data.}
We generate a simulation environment where we instantiate the necessary numbers of points and lines to test each configuration. To define a 3D line, we parameterize it as a tuple of two points. All these points are randomly generated inside an origin centered, axis aligned, 3D cube of edge size $0.6$, which represents the model's frame of reference. We then apply a random 3D transformation which guarantees that all origin of the model will lie somewhere in $[-0.5,0.5]\times[-0.5,0.5]\times[0.4, 2.0]$. To project the point onto the image plane, we adopt the same camera intrinsics of the Kinect v1. Ultimately, we are trying to replicate similar conditions to those present in the LINEMOD dataset \cite{linemod}, which contains scenes targeting object pose estimation tasks. We apply Gaussian noise of various levels to the projected pixels to simulate noise in the camera. With every single run, we instantiate new random elements, a new random pose and apply random pixel noise to the projections. We then submit all methods to the same realization to ensure that we have a direct comparison in every single run. This ensures that even if a specific realization is degenerate or simply challenging, all methods are subjected to it. We compare our method versus a number of other available approaches in the scenario of $4+$ points, $4+$ lines and the combination of $4+$ points and lines. We evaluate all methods for different levels of Gaussian pixel noise and number of elements in the scene. We benchmark against Vakhitov et al. \cite{vakhitov2016accurate} which developed EPnPL and OPnPL as extensions of the original EPnP \cite{pnp_lepetit2009epnp} and OPnP \cite{Zheng_2013_ICCV}, to support mixed combinations of points and lines. For points-only scenarios, the existing number of PnP methods in the literature is considerable, so we opted to focus on the more popular ones such as EPnP, UPnP \cite{pnp_kneip2014upnp} and OPnP\footnote{EPnP implementation from OpenCV \cite{opencv}, UPnP implementation from OpenGV \cite{opengv} and OPnP implementation from Vakhitov et al. \cite{vakhitov2016accurate}.}. For the lines-only scenarios, we compare against the works of Mirzaei and Roumeliotis \cite{mirzaei2011globally}, RPnL \cite{zhang2012robust}, EPnPL and OPnPL\footnote{All PnL implementations from Vakhitov et al. \cite{vakhitov2016accurate}.}. Finally, for the mixed scenarios we only compare directly with EPnPL and OPnPL. The condensed results for all these scenarios are displayed in Fig. \ref{fig:results-synth}. We can verify that our method achieves results in line with the most precise state of the art methods.

\begin{figure}
\centering
\begin{tabular}{cc}
\includegraphics[width=0.49\linewidth]{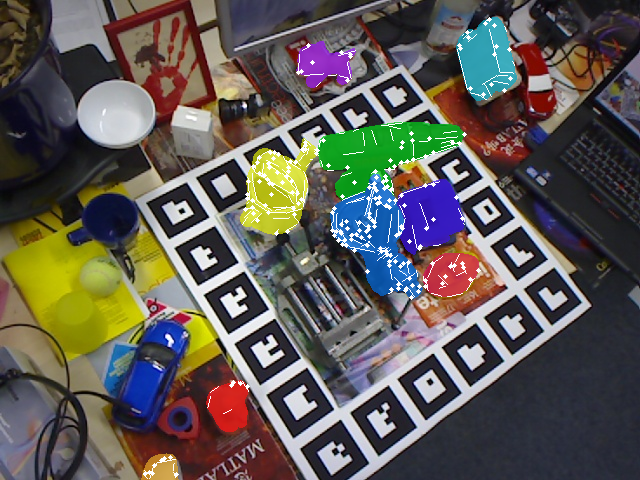}&
\includegraphics[width=0.49\linewidth]{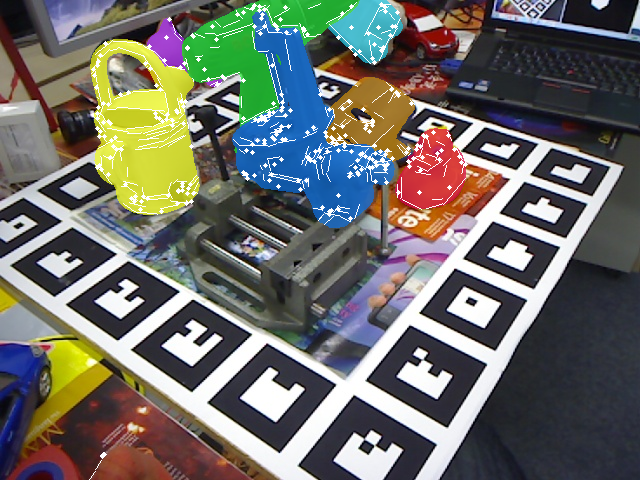}\\
(a)&(b)
\end{tabular}
\caption{Two views from the Occlusion dataset, showing the capability of our method to handle real scenes. The key points and lines used as correspondences are represented in white (figure best seen in colour).}
\label{fig:results-real}
\vspace{-0.5cm}
\end{figure}

\textbf{Real Data.}
To generate real data we make use of the richly annotated bench vise scene commonly named as the Occlusion dataset \cite{object-coords}. 
The dataset provides groundtruth 3D models and poses.
In order to establish ground truth 2D-3D correspondences, we borrow the strategy employed in \cite{object-coords} which allows inferring for each pixel belonging to an object, its normalized 3D object coordinate. The unnormalized 3D object coordinate is recovered from the lengths of the 3D bounding box dimensions of the object. Using the Occlusion dataset allows having a direct ground truth pose comparison and generate dense object masks. In order to select meaningful 2D points from the image, we use to SIFT \cite{sift} to establish the point proposals, pruning in a later stage any proposal which does not lie inside the ground-truth object masks. To extract line segment we use LSD \cite{von2010lsd} to generate segment proposals. These are also later pruned to regions belonging to the ground truth masks. We show our qualitative estimation results from points and lines in Figure \ref{fig:results-real}.
We are able to successfully retrieve meaningful poses. \footnote{The code implementation of CvxPnPL is available at \url{https://github.com/SergioRAgostinho/cvxpnpl}.}

\section{Conclusion}
\label{sec:conclusion}

We introduced the first convex approach to the central absolute problem from mixed point and line correspondences. We formulated our optimization problem as a QCQP and relaxed it into a SDP. We then verified the rank of the relaxed solution is equal to the number of ambiguous poses in the problem and derived approaches for retrieving all poses in situations containing up to 4 ambiguous solutions. We showed that our method is competitive with the best state-of-the-art algorithms under synthetic conditions and qualitatively validated its performance on a real dataset.

%
%
%
\bibliographystyle{splncs04}
\bibliography{submission}

\end{document}


\title{Supplementary Material for\\
\textit{CvxPnPL: A Unified Convex Solution to the Absolute Pose Estimation Problem from Point and Line Correspondences}}

\titlerunning{CvxPnPL - Supplementary Material}
%
\author{Sérgio Agostinho\inst{1}\orcidID{0000-0001-7008-1756} \and
João Gomes\inst{1}\orcidID{0000-0002-3524-5556} \and
Alessio Del Bue\inst{2}\orcidID{0000-0002-2262-4872}}
%
\authorrunning{S. Agostinho, J. Gomes and A. Del Bue}
%

\institute{
Instituto de Sistemas e Robótica\\
Instituto Superior Técnico, Universidade de Lisboa \\
Portugal\\
\email{sergio.agostinho@tecnico.ulisboa.pt,jpg@isr.tecnico.ulisboa.pt}\\
\and
Fondazione Istituto Italiano di Tecnologia\\
Genoa, Italy\\
\email{alessio.delbue@iit.it}\\
}
\maketitle              
%
%

\section{Introduction}

In this supplementary document, we present the full derivations and extra clarifications on: how to achieve the linear system of equations based on the geometric constraints of the Perspective-n-Points-and-Lines (PnPL); how to rewrite the original optimization problem in a Quadratically Constrained Quadratic Program (QCQP) friendly formulation; how we empirically verified that the rank of $\mat Z$ acts as a proxy to the number of solutions; how to reformulate the rotation constraint on the solution space in a system of second-order polynomial equations.
%
%
\section{Useful Vectorization Properties}

See \cite{schacke2004kronecker} for an insightful discussion on the following identities:
\begin{equation}
\tr(\mat{A}^\top \mat{B}) = \ovec(\mat{A})^\top \ovec(\mat{B})
\end{equation}
\begin{equation}
\ovec(\mat{A}\mat{X}\mat{B}) = (\mat{B}^\top \otimes \mat{A})\ovec(\mat{X}) \label{eq:vec-trace}
\end{equation}
\begin{equation}
\ovec(\vec{a}\vec{b}^\top) = \vec{b} \otimes \vec{a}
\end{equation}
\begin{equation}
\tr(\mat{A}^\top \mat{X}^\top \mat{B} \mat{Y}) = \ovec(\mat{X})^\top (\mat{A} \otimes \mat{B}) \ovec(\mat{Y}). \label{eq:vec-trace-quad}
\end{equation}

\section{Geometric Constraints}

\subsection{Point Correspondences}

\begin{figure}[t]
\begin{center}
\includegraphics[width=7.6cm]{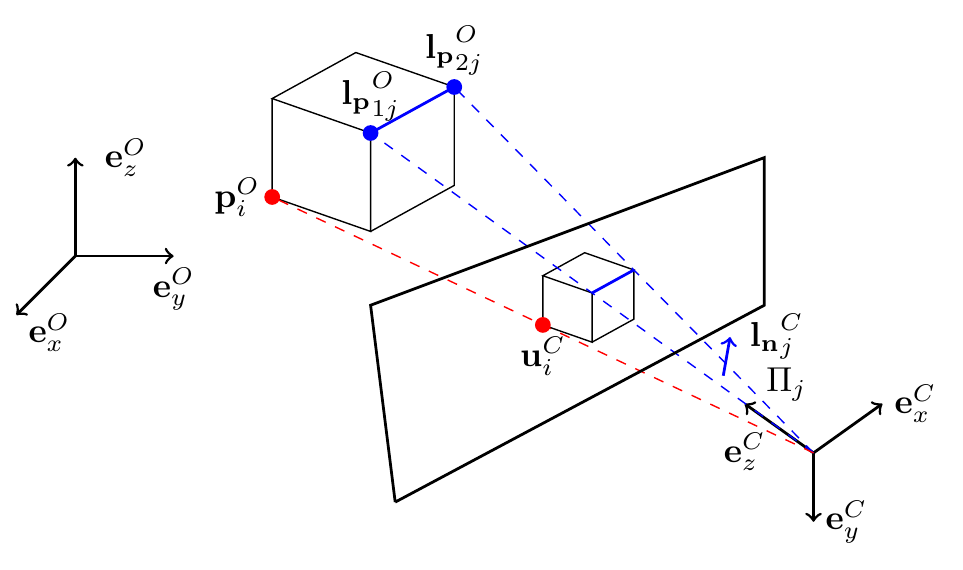}
\end{center}
\caption{Example of 2D-3D correspondences from a single point and line. Elements belonging to point correspondences are represented in red and elements belonging to line correspondences are represented in blue. The superscripts $O$ and $C$ denote that the element belongs to the object and camera's reference frames, respectively.}
\label{fig:correspondences}
\end{figure}
Consider $\vec{u}_i$ as the bearing vector associated with the 3D point $\vec{p}_i$ (refer to Figure \ref{fig:correspondences}). The transformation from the model/object space to the camera is parameterized by the rotation matrix $\mat{R}$ and the translation vector $\vec{t}$. The collinearity constraint of a point allows to us write,
\begin{equation}
\lfloor \vec{u}_i \rfloor_\times \left(\mat{R} \mathbf{p}_i + \mathbf{t}\right) = 0, \label{eq:skew}
\end{equation}
where $\lfloor \vec{u} \rfloor_\times$ is the skew symmetric matrix representation of the 3D vector $\vec{u}$
\begin{equation}
\lfloor \vec{u} \rfloor_\times = \left[ \begin{matrix}
0 & -u_z & u_y \\
u_z & 0 & -u_x \\
-u_y & u_x & 0
\end{matrix} \right].
\end{equation}
We can represent Eq. \eqref{eq:skew} isolating the vectorized representation of $\mat{R}$,
\begin{equation}
\vec{r} = \ovec(\mat{R}),
\end{equation}
as
\begin{eqnarray}
\lfloor \vec{u}_i \rfloor_\times \left(\mat{R} \mathbf{p}_i + \mathbf{t}\right) &=& 0\\
\lfloor \vec{u}_i \rfloor_\times \mat{R} \mathbf{p}_i + \lfloor \vec{u}_i \rfloor_\times \mathbf{t} &=& 0 \quad \text{: (distributive property)} \\
\underbrace{(\mathbf{p}^\top_i \otimes \lfloor \vec{u}_i \rfloor_\times)}_{\mat{C_p}_i} \vec{r} + \underbrace{\lfloor \vec{u}_i \rfloor_\times}_{\mat{N_p}_i} \mathbf{t} &=& 0. \quad \text{: (applying \eqref{eq:vec-trace}})
\end{eqnarray}
Matrices $\mat{C_p}_i$ and $\mat{N_p}_i$ are sizes $3\times9$ and $3\times2$, respectively. Each point correspondence contributes with 3 equations and upon stacking all $n$ points yields,
\begin{eqnarray}
\mat{C_p} &=& \left[\begin{matrix}\mat{C_p}_1 \\ \vdots \\ \mat{C_p}_n\end{matrix}\right] \\
\mat{N_p} &=& \left[\begin{matrix}\mat{N_p}_1 \\ \vdots \\ \mat{N_p}_n\end{matrix}\right],
\end{eqnarray}
which are $3n\times9$ and $3n\times3$ sized matrices, respectively. This ultimately results in the homogeneous system of equations
\begin{equation}
\mat{C_p} \vec{r} + \mat{N_p} \vec{t} = 0.
\end{equation}

\subsection{Line Correspondences}

Consider a 3D line defined by two points, which we represent by the tuple $(\vec{l_p}_{1j}, \vec{l_p}_{2j})$. Also consider $\vec{l_n}_j$, the normal to the plane formed between the 3D line and the origin of the camera.
The coplanarity constraints of the line, define
\begin{equation}
\vec{l_n}^\top_j (\mat{R} \vec{l_p}_{ij} + \vec{t}) = 0 \quad : i = {1, 2}. \label{eq:line-const-pt}
\end{equation}
From Eq. \eqref{eq:line-const-pt}, we have
\begin{eqnarray}
\vec{l_n}^\top_j (\mat{R} \vec{l_p}_{ij} + \vec{t}) &=& 0 \\
\vec{l_n}^\top_j\mat{R} \vec{l_p}_{ij} + \vec{l_n}^\top_j\vec{t} &=& 0 \quad \text{: (distributive property)} \\
(\vec{l_p}_{ij}^\top \otimes \vec{l_n}^\top_j)\vec{r} + \vec{l_n}^\top_j\vec{t} &=& 0 \quad \text{: (applying \eqref{eq:vec-trace})} \\
{\underbrace{(\vec{l_p}_{ij} \otimes \vec{l_n}_j)}_{\vec{c_l}_{ij}}}^\top \vec{r} + \vec{l_n}^\top_j\vec{t} &=& 0. \quad \text{: (single out ${}^\top$)}
\end{eqnarray}
The vector $\vec{c_l}_{ij} \in \R^9$. Stacking the contributions from all $m$ lines yields
\begin{eqnarray}
\mat{C_l} &=& \left[\begin{matrix}\vec{c_l}^\top_{11} \\ \vec{c_l}^\top_{21} \\ \vdots \\ \vec{c_l}^\top_{1m} \\ \vec{c_l}^\top_{2m}\end{matrix}\right] \\
\mat{N_l} &=& \left[\begin{matrix}\vec{l_n}^\top_1 \\ \vec{l_n}^\top_1 \\ \vdots \\ \vec{l_n}^\top_m \\ \vec{l_n}^\top_m \end{matrix}\right],
\end{eqnarray}
which are matrices of size $2m\times9 $ and $2m\times3$, respectively. The full system of equations assumes the form
\begin{equation}
\mat{C_l} \vec{r} + \mat{N_l} \vec{t} = 0.
\end{equation}

\section{The QCQP Reformulation}

In the main paper, we reach the conclusion that our optimization problem can be described by the following Quadratically Constrainted Quadratic Program (QCQP),
\begin{eqnarray}
\operatorname*{min}_{\mathbf{r}} & & \|\mat{A} \mathbf{r}\|^2  \\
\operatorname{s.t.\,} & & \mat{R}^\top \mat{R} = \mat{I}_3 \label{eq:constraint-cols}\\
& & \mat{R} \mat{R}^\top = \mat{I}_3 \label{eq:constraint-rows}\\
& & \mat{R}^{(i)} \times \mat{R}^{(j)} = \mat{R}^{(k)} \, : (i, j, k) \in \{(1, 2, 3), (2, 3, 1), (3, 1, 2)\}, \label{eq:constraint-right}
\end{eqnarray}
with $\mat{R}^{(i)}$ denoting the $i$-th column of $\mat{R}$. $A$ is a $(3n + 2m)\times 9$ matrix. However, we are still required to express it in a more canonical form, which conventional convex solvers support. In the next sections, we address this reformulation for each element of the problem. Out of convenience, we resort to the homogenized representation of $\vec{r}$,
\begin{equation}
\tilde{\vec{r}} = \left[\begin{matrix}\vec{r} \\ 1\end{matrix}\right],
\end{equation}
as it eases bundling all constant and linear terms with respect to $\vec{r}$, into a quadratic form with respect to $\tilde{\vec{r}}$.

\subsection{Cost Function}

The cost function is defined by
\begin{eqnarray}
\| A \vec{r} \|^2 &=&(A \vec{r})^\top (A \vec{r}) \quad : \text{($\|\vec{v}\|^2 = \vec{v}^\top \vec{v}$ for $\vec{v} \in \mathbb{R}^n$)} \\
&=& \vec{r}^\top A^\top A \vec{r} \quad : \text{(apply transpose of product properties)} \\
&=& \tilde{\vec{r}}^\top \underbrace{\left[\begin{matrix}A^\top A & 0_{9\times1} \\ 0_{1\times9} & 0\end{matrix}\right]}_{Q_0} \tilde{\vec{r}}. \quad : \text{(replace $\vec{r}$ for $\tilde{\vec{r}}$)}
\end{eqnarray}

\subsection{Orthogonality of Rows}
\label{sec:ortho-rows}

The constraint associated with the orthogonality of rows, provides 6 linearly independent equations,
\begin{equation}
\mat{R}\mat{R}^\top = \mat{I}_3, \label{eq:ortho-rows}
\end{equation}
where the matrix $\mat{I}_3$ represents the identity of size $3\times3$. We introduce the unit vector $\vec{e}_i \in \mathbb{R}^3$, whose component $i$ is set to 1 and the remainder to 0, as well as the $3\times3$ matrix
\begin{equation}
\mat{E}_{ij} = \vec{e}_i \vec{e}_j^\top,
\end{equation}
which is composed of a unique element 1 at the row $i$ and column $j$, while all the remainder are also 0. Equipped with these new definitions, we are able express the constraint in \eqref{eq:ortho-rows}, with respect to each individual component. In general, for a given matrix $\mat{A} \in \mathbb{R}^{3\times3}$ we have that:
\begin{equation}
\vec{e}_i^\top \mat{A} \vec{e}_j = a_{ij}.
\end{equation}
Employing the same property we have
\begin{eqnarray}
\vec{e}_i^\top (\mat{R}\mat{R}^\top - \mat{I}_3)\vec{e}_j &=& 0 \\
\vec{e}_i^\top \mat{R}\mat{R}^\top \vec{e}_j - \vec{e}_i^\top \mat{I}_3\vec{e}_j &=& 0 \quad : \text{(distributive property)}\\
\vec{e}_i^\top \mat{R}\mat{R}^\top \vec{e}_j - \delta_{ij} &=& 0 \quad : \text{(Kronecker delta)}\\
\tr(\vec{e}_i^\top \mat{R}\mat{R}^\top \vec{e}_j) - \delta_{ij} &=& 0 \quad : \text{(trace of a scalar)}\\
\tr(\mat{R}^\top \vec{e}_j\vec{e}_i^\top \mat{R}) - \delta_{ij} &=& 0 \quad : \text{(cyclic property of the trace)}\\
\tr(\mat{R}^\top \mat{E}_{ji}\mat{R}) - \delta_{ij} &=& 0 \quad : \text{(substitution of $\mat{E}_{ji}$)}\\
\tr(\mat{I}_3\mat{R}^\top \mat{E}_{ji} \mat{R}) - \delta_{ij} &=& 0 \quad : \text{(identity matrix)}\\
\vec{r}^\top (\mat{I}_3 \otimes \mat{E}_{ji}) \vec{r} - \delta_{ij} &=& 0 \quad : \text{(applying \eqref{eq:vec-trace-quad})}\\
\tilde{\vec{r}}^\top \underbrace{\left[\begin{matrix}\mat{I}_3 \otimes \mat{E}_{ji} & 0_{9\times1}\\
0_{1\times9} & - \delta_{ij}\end{matrix}\right]}_{\mat{Q_r}_{ij}} \tilde{\vec{r}} &=& 0. \quad : \text{(with respect to $\tilde{\vec{r}}$)}
\end{eqnarray}
Iterating for all indexes, we compose 6 constraints:
\begin{equation}
\tilde{\vec{r}}^\top \mat{Q_r}_{ij} \tilde{\vec{r}} = 0 \quad : i = 1, \dots, 3; j = i, \dots, 3.
\end{equation}

\subsection{Orthogonality of Columns}

The derivation is very similar to Sec. \ref{sec:ortho-rows}. This time we start from the following constraint:
\begin{eqnarray}
\vec{e}_i^\top (\mat{R}^\top \mat{R} - \mat{I}_3)\vec{e}_j &=& 0 \\
\vec{e}_i^\top \mat{R}^\top \mat{R} \vec{e}_j - \vec{e}_i^\top \mat{I}_3\vec{e}_j &=& 0 \quad : \text{(distributive property)}\\
\vec{e}_i^\top \mat{R}^\top \mat{R} \vec{e}_j - \delta_{ij} &=& 0 \quad : \text{(Kronecker delta)}\\
\tr(\vec{e}_i^\top \mat{R}^\top \mat{R} \vec{e}_j) - \delta_{ij} &=& 0 \quad : \text{(trace of a scalar)}\\
\tr(\vec{e}_j\vec{e}_i^\top \mat{R}^\top \mat{R}) - \delta_{ij} &=& 0 \quad : \text{(cyclic property of the trace)}\\
\tr(\mat{E}_{ji} \mat{R}^\top \mat{R}) - \delta_{ij} &=& 0 \quad : \text{(substitution of $\mat{E}_{ji}$)}\\
\tr(\mat{E}_{ij}^\top\mat{R}^\top \mat{R}) - \delta_{ij} &=& 0 \quad : \text{(transpose $\mat{E}_{ji}$)}\\
\tr(\mat{E}_{ij}^\top\mat{R}^\top \mat{I}_3 \mat{R}) - \delta_{ij} &=& 0 \quad : \text{(identity matrix)}\\
\vec{r}^\top (\mat{E}_{ij} \otimes \mat{I}_3) \vec{r} - \delta_{ij} &=& 0 \quad : \text{(applying \eqref{eq:vec-trace-quad})}\\
\tilde{\vec{r}}^\top \underbrace{\left[\begin{matrix}\mat{E}_{ij} \otimes \mat{I}_3 & 0_{9\times1}\\
0_{1\times9} & - \delta_{ij}\end{matrix}\right]}_{\mat{Q_c}_{ij}} \tilde{\vec{r}} &=& 0. \quad : \text{(with respect to $\tilde{\vec{r}}$)}
\end{eqnarray}
Iterating for all indexes, composes 6 constraints as:
\begin{equation}
\tilde{\vec{r}}^\top \mat{Q_c}_{ij} \tilde{\vec{r}} = 0 \quad : i = 1, \dots, 3; j = i, \dots, 3.
\end{equation}

\subsection{Determinant - Right-Hand Convention}

We start from the original constraints as defined in Eq. \eqref{eq:constraint-right}:
\begin{eqnarray}
\mat{R}^{(i)} \times \mat{R}^{(j)} = \mat{R}^{(k)}, \quad : (i, j, k) \in \{(1, 2, 3), (2, 3, 1), (3, 1, 2)\}
\end{eqnarray}
where $\mat{R}^{(i)}$, represents the $i$-th column of $\mat{R}$. These are a total 9 constraints and once more we define them with respect to its individual components. Starting from
\begin{eqnarray}
\vec{e}_l^\top (\mat{R}^{(i)} \times \mat{R}^{(j)} - \mat{R}^{(k)}) &=& 0 \\
\vec{e}_l^\top (\mat{R}^{(i)} \times \mat{R}^{(j)}) - \vec{e}_l^\top \mat{R}^{(k)} &=& 0 \quad : \text{(distributive property)}\\
{\mat{R}^{(j)}}^\top (\vec{e}_l \times \mat{R}^{(i)}) - \vec{e}_l^\top \mat{R}^{(k)} &=& 0 \quad : (\vec{a}\cdot (\vec{b} \times \vec{c}) = \vec{c}\cdot (\vec{a} \times \vec{b}))\\
{\mat{R}^{(j)}}^\top \skewm{\vec{e}_l} \mat{R}^{(i)} - \vec{e}_l^\top \mat{R}^{(k)} &=& 0 \quad : \text{(skew symmetric matrix)}\\
\vec{e}_j^\top \mat{R}^\top \skewm{\vec{e}_l} \mat{R}\vec{e}_i - \vec{e}_l^\top \mat{R}\vec{e}_k &=& 0 \quad : (\mat{R}^{(i)} = \mat{R}\vec{e}_i)\\
\vec{e}_j^\top \mat{R}^\top \skewm{\vec{e}_l} \mat{R}\vec{e}_i - (\vec{e}_k \otimes \vec{e}_l)^\top \vec{r} &=& 0 \quad \text{: (applying \eqref{eq:vec-trace})} \\
\tr(\vec{e}_j^\top \mat{R}^\top \skewm{\vec{e}_l} \mat{R}\vec{e}_i) - (\vec{e}_k \otimes \vec{e}_l)^\top \vec{r} &=& 0 \quad \text{: (trace of the scalar)}\\
\tr(\vec{e}_i\vec{e}_j^\top \mat{R}^\top \skewm{\vec{e}_l} \mat{R}) - (\vec{e}_k \otimes \vec{e}_l)^\top \vec{r} &=& 0 \quad \text{: (cyclic property of the trace)}\\
\tr(\mat{E}_{ji}^\top \mat{R}^\top \skewm{\vec{e}_l} \mat{R}) - (\vec{e}_k \otimes \vec{e}_l)^\top \vec{r} &=& 0 \quad \text{: (substitution of $\mat{E}_{ji}$)} \\
\tr(\mat{E}_{ji}^\top \mat{R}^\top \skewm{\vec{e}_l} \mat{R}) - \vec{r}^\top (\vec{e}_k \otimes \vec{e}_l) &=& 0 \quad \text{: (scalar transpose)}\\
\vec{r}^\top (\mat{E}_{ji} \otimes \skewm{\vec{e}_l}) \vec{r} - \vec{r}^\top (\vec{e}_k \otimes \vec{e}_l) &=& 0 \quad \text{: (applying \eqref{eq:vec-trace-quad})}
\end{eqnarray}
\begin{eqnarray}
\tilde{\vec{r}}^\top \underbrace{\left[\begin{matrix} \mat{E}_{ji} \otimes \skewm{\vec{e}_l} &  - (\vec{e}_k \otimes \vec{e}_l) \\
0_{1\times9} & 0 \end{matrix}\right]}_{\mat{Q_d}_{(i,j,k)}} \tilde{\vec{r}}^\top &=& 0. \quad \text{: (with respect to $\tilde{\vec{r}}$)}
\end{eqnarray}
Iterating over all indexes, composes 9 constraints
\begin{equation}
\tilde{\vec{r}}^\top \mat{Q_d}_{(i,j,k)} \tilde{\vec{r}} = 0 \quad : (i, j, k) \in \{(1, 2, 3), (2, 3, 1), (3, 1, 2)\}.
\end{equation}

\section{Rank of $\mat Z$ as an Indicator of the Number of Solutions}

One of the claims made in the main paper, is of the rank of $\mat Z$ being equal to the number of minima of the non-convex optimization problem i.e., prior to the rank relaxation. The intuition for this claim came first from the reports in \cite{Briales_2018_CVPR}, where the authors observed that the relative pose problem has 4 ambiguous solutions, so for this reason the rank of their relaxed solution was 4.

This is also valid, even with a different problem and formalisation, for our proposed solution to the Perspective-3-Points case -- which has also 4 solutions. In such case, we were able to verify the same rank-4 property on the solution matrix.
The common, non-degenerate, non-minimal PnP case as a unique solution. Most PnP methods perform reliably well under these conditions and we managed to verify here as well the rank of $Z$ being consistently 1.

At last, among the cases we provide a solution for, the rank 2 was the only one remaining case to analyse. Generating 2D-3D correspondences which generate a rank 2 matrix $\mat Z$ is fairly common under certain particular configurations of lines 
for the Perspective-n-Lines problem with 4 lines. As a validation procedure, we looked at the solutions provided by the OPnPL method \cite{vakhitov2016accurate} and evaluated them on our cost function. Despite OPnPL returning more solutions than required and some being occasionally duplicate, we could verify that also here, the number of minima agrees with the rank of the $\mat Z$ we obtain in our method. This further validates that the rank of $\mat Z$ can be used as a proxy for the number of solutions in CvxPnPL.

\section{Composing the Linear System of Quadrics}

One of the claims made in the paper, states that under minimal configurations, the solution space of all admissible solutions, is given by the linear decomposition
\begin{equation}
\vec{r} = \sum_{k=1}^{K-1} \alpha_k' \mathbf v_k' + \mathbf v_0',\label{eq:linear-comb-vec}
\end{equation}
Where $K$ is the rank of matrix $\mat Z$. Consider now the case $K=4$, which is the highest rank we address in the paper. The three unknowns $\alpha_1'$, $\alpha_2'$ and $\alpha_3$ will be designated by the letters $a$, $b$ and $c$ and we will drop the $'$ superscript on the vectors for convenience. We resort to the inverse of the vectorizing operator $\ovec^{-1}$ to reformulate Eq. \eqref{eq:linear-comb-vec}. Defining
\begin{equation}
\mat{V}_i = \ovec^{-1}(\vec{v}_i) : i = 0, \dots, 3,
\end{equation}
we can rewrite it as
\begin{equation}
\mat{R} = a \mat{V}_1 + b\mat{V}_2 + c\mat{V}_3 + \mat{V}_4.
\end{equation}
Once more, for $\mat{R}$ to be valid, it needs to respect the same set of constraints
\begin{eqnarray}
\mat{R}\mat{R}^\top &=& \mat{I}_3 \\
\mat{R}^\top \mat{R} &=& \mat{I}_3 \\
\mat{R}^{(i)} \times \mat{R}^{(j)} &=& \mat{R}^{(k)} \quad : (i, j, k) \in \{(1, 2, 3), (2, 3, 1), (3, 1, 2)\}.
\end{eqnarray}
In the next sections we will show how to rewrite this combined set of 21 constraints, into the linear system
\begin{equation}
\mat{A} \left[\begin{matrix}a^2 \\ b^2 \\ c^2 \\ ab \\ac \\ bc \\ a \\ b \\c \\ 1\end{matrix}\right] = 0, \label{eq:final-linear}
\end{equation}
where $\mat{A}$ is a matrix of size $21\times10$.

\subsection{Reformulating the Quadratic System of Equations into a Linear System with Respect to Quadratic Terms}
\label{sec:quad-2-lin}

One important aspects of this formulation is recognizing how to convert the quadratic system to its ``linear" form, with respect to quadratic terms. Consider the vector
\begin{equation}
\vec{v} = \left[\begin{matrix}a\\ b \\c \\1\end{matrix}\right],
\end{equation}
and the quadratic expression
\begin{equation}
\vec{v}^\top \mat{P} \vec{v} = 0, \label{eq:quad-gen}
\end{equation}
where $\mat{P} \in \mathbb{R}^{4\times 4}$. We can reformulate Eq. \eqref{eq:quad-gen} to its ``linear'' form as
\begin{equation}
\left[\begin{matrix}
P_{11} \\
P_{22} \\
P_{33} \\
P_{12} + P_{21} \\
P_{13} + P_{31} \\
P_{23} + P_{32} \\
P_{14} + P_{41} \\
P_{24} + P_{42} \\
P_{34} + P_{43} \\
P_{44}
\end{matrix}\right]^\top \left[\begin{matrix}a^2 \\ b^2 \\ c^2 \\ ab \\ac \\ bc \\ a \\ b \\c \\ 1\end{matrix}\right] = 0
\end{equation}
In the next sections we will focus once more on describing all constraints in their natural quadratic form, always with the outlook that the previous reformulation can be applied and that each quadratic constraint will contribute with a row in the final linear system in Eq. \eqref{eq:final-linear}.

\subsection{Orthogonality of Columns}

We start by writing Eq. \eqref{eq:linear-comb-vec} in a more compact linear form
\begin{equation}
\vec{r} = \mat{V} \vec{q}, \label{eq:r-compact}
\end{equation}
where
\begin{eqnarray}
\mat{V} &=& \left[\begin{matrix}\vec{v}_1 & \vec{v}_2 & \vec{v}_3 & \vec{v}_0 \end{matrix}\right] \\
\vec{q} &=& \left[\begin{matrix}a \\ b \\ c \\ 1\end{matrix}\right].
\end{eqnarray}
The first important step is to define an operation which allows us to select each column of $\mat{R}$. Using the index $i$ to designate the desired column, we can write
\begin{equation}
\vec{r_c}_i = (\vec{e}_i \otimes \mat{I}_3)^\top \vec{r}. \label{eq:col-selector}
\end{equation}
With this mechanism in place, the orthogonality of the columns specifies that
\begin{equation}
\vec{r_c}_i^\top \vec{r_c}_j = \delta_{ij} \quad \text{for} \quad i = \{1, 2, 3\}, j = \{i, \dots, 3\}.
\end{equation}
Substituting the appropriate terms
\begin{eqnarray}
\vec{r_c}_i^\top \vec{r_c}_j - \delta_{ij} &=& 0 \\
\vec{r}^\top (\vec{e}_i \otimes \mat{I}_3) (\vec{e}_j \otimes \mat{I}_3)^\top \vec{r} - \delta_{ij} &=& 0 \quad \text{: (substituting \eqref{eq:col-selector})}\\
\vec{q}^\top \mat{V}^\top (\vec{e}_i \otimes \mat{I}_3) (\vec{e}_j \otimes \mat{I}_3)^\top \mat{V} \vec{q} - \delta_{ij} &=& 0 \quad \text{: (substituting \eqref{eq:r-compact})}\\
\vec{q}^\top \mat{V}^\top (\vec{e}_i \otimes \mat{I}_3) (\vec{e}_j \otimes \mat{I}_3)^\top \mat{V}\vec{q} & & \\
- \vec{q}^\top \left[\begin{matrix}0_{3\times3} & 0_{3\times1}\\ 0_{1\times3} & \delta_{ij}\end{matrix}\right]\vec{q} &=& 0. \quad \text{: (representing $\delta_{ij}$ in terms of $\vec{q}$)}
\end{eqnarray}
Considering all indices
\begin{eqnarray}
\mat{P_c}_{ij} &=&  \mat{V}^\top (\vec{e}_i \otimes \mat{I}_3) (\vec{e}_j \otimes \mat{I}_3)^\top \mat{V} - \left[\begin{matrix}0_{3\times3} & 0_{3\times1}\\ 0_{1\times3} & \delta_{ij}\end{matrix}\right] \\
\vec{q}^\top \mat{P_c}_{ij}\vec{q} &=& 0 \quad \text{for} \quad i = \{1, 2, 3\}, j = \{i, \dots, 3\},
\end{eqnarray}
it contributes 6 equations to the system in Eq. \eqref{eq:final-linear}.

\subsection{Orthogonality of Rows}

Similarly to the previous section, we start by defining a selector operator which allows us to isolate the rows of $\mat{R}$ from $\vec{r}$. Not surprisingly, we can verify that such operator can be built from commuting both terms in the Kronecker product.
\begin{equation}
\vec{r_r}_i = (\mat{I}_3 \otimes \vec{e}_i)^\top \vec{r}. \label{eq:row-selector}
\end{equation}
With this operator in place, the orthogonality of the rows specifies that
\begin{equation}
\vec{r_r}_i^\top \vec{r_r}_j = \delta_{ij} \quad \text{for} \quad i = \{1, 2, 3\}, j = \{i, \dots, 3\}.
\end{equation}
Substituting the appropriate terms
\begin{eqnarray}
\vec{r_r}_i^\top \vec{r_r}_j - \delta_{ij} &=& 0 \\
\vec{r}^\top (\mat{I}_3 \otimes \vec{e}_i) (\mat{I}_3 \otimes \vec{e}_i)^\top \vec{r} - \delta_{ij} &=& 0 \quad \text{: (substituting \eqref{eq:row-selector})}\\
\vec{q}^\top \mat{V}^\top (\mat{I}_3 \otimes \vec{e}_i) (\mat{I}_3 \otimes \vec{e}_i)^\top \mat{V} \vec{q} - \delta_{ij} &=& 0 \quad \text{: (substituting \eqref{eq:r-compact})}\\
\vec{q}^\top \mat{V}^\top (\mat{I}_3 \otimes \vec{e}_i) (\mat{I}_3 \otimes \vec{e}_i)^\top \mat{V}\vec{q} & & \\
- \vec{q}^\top \left[\begin{matrix}0_{3\times3} & 0_{3\times1}\\ 0_{1\times3} & \delta_{ij}\end{matrix}\right]\vec{q} &=& 0. \quad \text{: (representing $\delta_{ij}$ in terms of $\vec{q}$)}
\end{eqnarray}

Considering all indices
\begin{eqnarray}
\mat{P_r}_{ij} &=& \mat{V}^\top (\mat{I}_3 \otimes \vec{e}_i) (\mat{I}_3 \otimes \vec{e}_i)^\top \mat{V} - \left[\begin{matrix}0_{3\times3} & 0_{3\times1}\\ 0_{1\times3} & \delta_{ij}\end{matrix}\right]  \\
\vec{q}^\top \mat{P_r}_{ij}\vec{q} &=& 0 \quad \text{for} \quad i = \{1, 2, 3\}, j = \{i, \dots, 3\},
\end{eqnarray}
it contributes 6 additional equations to the system in Eq. \eqref{eq:final-linear}.

\subsection{Determinant - Right Hand Convention}

Reusing the column selector operator from Eq. \eqref{eq:col-selector}, the right-hand convention specifies that
\begin{equation}
\vec{r_c}_i \times \vec{r_c}_j = \vec{r_c}_k \quad \text{for} \quad (i, j, k) \in \{(1, 2, 3), (2, 3, 1), (3, 1, 2)\}.
\end{equation}
We resort to the unit vector $\vec{e}_l$, to specify the constraint with respect to each individual component. Starting with
\begin{eqnarray}
\vec{e}_l^\top (\vec{r_c}_i \times \vec{r_c}_j - \vec{r_c}_k) &=& 0 \\
\vec{e}_l^\top (\vec{r_c}_i \times \vec{r_c}_j) - \vec{e}_l^\top\vec{r_c}_k &=& 0 \quad \text{: (distributive property)}\\
\vec{r_c}_j^\top (\vec{e}_l \times \vec{r_c}_i) - \vec{e}_l^\top\vec{r_c}_k &=& 0 \quad : (\vec{a}\cdot (\vec{b} \times \vec{c}) = \vec{c}\cdot (\vec{a} \times \vec{b}))\\
\vec{r_c}_j^\top \skewm{\vec{e}_l} \vec{r_c}_i - \vec{e}_l^\top\vec{r_c}_k &=& 0 \quad \text{: (skew symmetric matrix)}\\
\vec{r}^\top (\vec{e}_j \otimes \mat{I}_3) \skewm{\vec{e}_l} (\vec{e}_i \otimes \mat{I}_3)^\top \vec{r} & & \\
- \vec{e}_l^\top (\vec{e}_k \otimes \mat{I}_3)^\top \vec{r} &=& 0 \quad \text{: (substituting \eqref{eq:col-selector})}\\
\vec{q}^\top \mat{V}^\top(\vec{e}_j \otimes \mat{I}_3) \skewm{\vec{e}_l} (\vec{e}_i \otimes \mat{I}_3)^\top \mat{V} \vec{q} & & \\
- \vec{e}_l^\top (\vec{e}_k \otimes \mat{I}_3)^\top \mat{V} \vec{q} &=& 0 \quad \text{: (substituting \eqref{eq:r-compact})}\\
\vec{q}^\top \mat{V}^\top(\vec{e}_j \otimes \mat{I}_3) \skewm{\vec{e}_l} (\vec{e}_i \otimes \mat{I}_3)^\top \mat{V}\vec{q} & & \\
- \vec{q}^\top \left[\begin{matrix}0_{3\times3} & 0_{3\times1}\\ \vec{e}_l^\top (\vec{e}_k \otimes \mat{I}_3)^\top & 0\end{matrix}\right]\vec{q} &=& 0. \quad \text{: (quadratic form of $\vec{q}$)}
\end{eqnarray}
Considering all indices we have
\begin{equation}
    \mat{P_d}_{ijkl} = \mat{V}^\top(\vec{e}_j \otimes \mat{I}_3) \skewm{\vec{e}_l} (\vec{e}_i \otimes \mat{I}_3)^\top \mat{V} - \left[\begin{matrix}0_{3\times3} & 0_{3\times1}\\ \vec{e}_l^\top (\vec{e}_k \otimes \mat{I}_3)^\top & 0\end{matrix}\right]
\end{equation}
\begin{eqnarray}
\vec{q}^\top \mat{P_d}_{ijkl}\vec{q} = 0 &\text{for}& (i, j, k) = \{(1, 2, 3), (2, 3, 1), (3, 1, 2)\},\\
& & l = \{1, 2, 3\}, \\
\end{eqnarray}
a total of 9 equations.

%
%
%
\bibliographystyle{splncs04}
\bibliography{supplemental}